\documentclass{article}
\usepackage{spconf,amsmath,epsfig}
\usepackage{array}
\usepackage{mdwmath}
\usepackage{mdwtab}
\usepackage{eqparbox}
\usepackage{url}
\usepackage{multirow}
\usepackage{bbding}
\usepackage{amsmath}
\usepackage{bm}
\usepackage{color}
\usepackage{array}
\usepackage{mdwmath}
\usepackage{mdwtab}
\usepackage{eqparbox}
\usepackage{url}
\usepackage{multirow}
\usepackage{bbding}
\usepackage{amsmath}
\usepackage {tabularx}
\usepackage{booktabs}

\let\OLDthebibliography\thebibliography
\renewcommand\thebibliography[1]{
  \OLDthebibliography{#1}
  \setlength{\parskip}{0pt}
  \setlength{\itemsep}{0pt plus 0.3ex}
}

\begin{document}

\def\x{{\mathbf x}}
\def\L{{\cal L}}

\title{Vision Transformer based Random Walk for Group Re-Identification}
%

\name{Guoqing Zhang$^{1}$, Tianqi Liu$^{1}$, Wenxuan Fang$^{1}$ and Yuhui Zheng$^{2\ast}$\thanks{$^{\ast}$Corresponding author. This work was partly supported by the National Natural Science Foundation of China (Grant No.61806099, U20B2065, U22B2056) and Natural Science Foundation of Jiangsu Province of China (Grant No.BK20211539 and BK20220107)}}
\address{%
$^{1}$ School of Computer Science, Nanjing University of Information Science and Technology, Nanjing, China\\
Email: guoqingzhang@nuist.edu.cn, liutianqi@nuist.edu.cn, fwenxuancontact@163.com \\
$^{2}$ The State Key Laboratory of Tibetan Intelligent Information Processing and Application, Qinghai\\Normal University, Xining, China. Email: zhengyh@vip.126.com\\
}

\maketitle

\begin{abstract}
Group re-identification (re-ID) aims to match groups with the same people under different cameras, mainly involves the challenges of group members and layout changes well. Most existing methods usually use the k-nearest neighbor algorithm to update node features to consider changes in group membership, but these methods cannot solve the problem of group layout changes. To this end, we propose a novel vision transformer based random walk framework for group re-ID. Specifically, we design a vision transformer based on a monocular depth estimation algorithm to construct a graph through the average depth value of pedestrian features to fully consider the impact of camera distance on group members relationships. In addition, we propose a random walk module to reconstruct the graph by calculating affinity scores between target and gallery images to remove pedestrians who do not belong to the current group. Experimental results show that our framework is superior to most methods.
\end{abstract}
%


\section{Introduction}
Person re-identification (re-ID) is a technique to re-identify the people under different cameras. Given a probe image, the task of person re-ID requires identifying the same person images from gallery images \cite{wang2020effective}, \cite{zhang2023multi}. In recent years, person re-ID has produced many derivative tasks in different directions, including clothes changing, visible-infrared and unsupervised person re-ID etc. However, the existing methods have mainly focused on single person re-ID, and people always walk in groups \cite{cai2010matching}, \cite{zhu2016consistent}, \cite{zheng2009associating}, \cite{zhu2020group}. Therefore, it is also crucial for group re-ID in real life.

\begin{figure}[!t]
\centering
\includegraphics[scale=0.35]{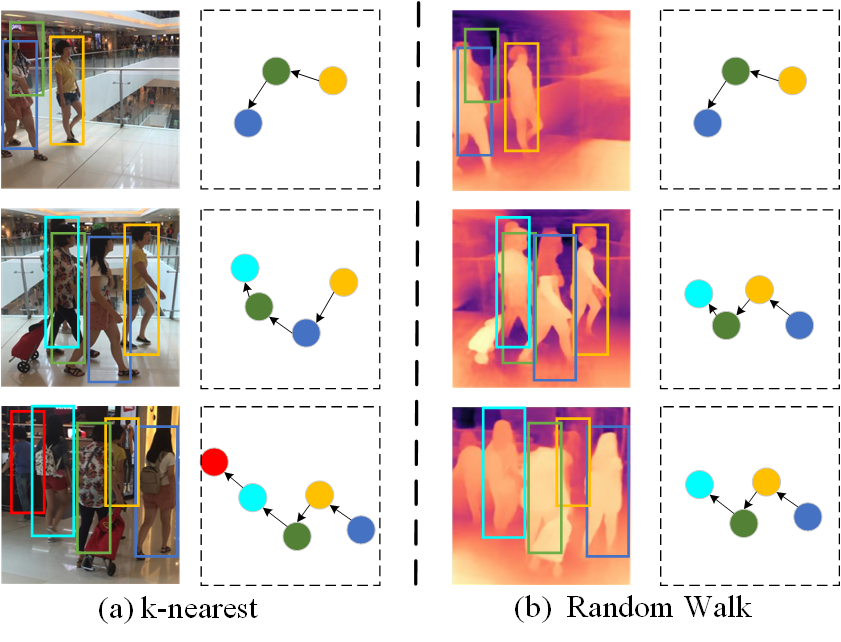}
\caption{Illustration two methods of constructing the graph: (a) the k-nearest neighbor algorithm and (b) the Random Walk method, which can effectively remove pedestrians (red nodes) that do not belong to the group.}
\label{fig:1}
\end{figure}

In comparison to person re-ID, group re-ID aims to re-identify group images with the same group members under different cameras and has additional challenges in addition to viewpoint changes and human pose changes: (i) Group layout changes: The layout of group members in a group is largely affected by different camera views constraints. Due to the dynamic movement of people, the relative positions of group members in a group can differ significantly in two camera views. (ii) Group members changes: Group members may dynamically join or leave a group frequently. The most existing methods for solving group re-ID mainly construct graphs through the k-nearest neighbor algorithm to transfer the group context information of adjacent members, enhancing the relationship between group members, as shown in Fig.~\ref{fig:1} (a). However, these methods not only ignore the impact of camera distance on group membership relationships, but also cannot fundamentally solve the problem of group layout changes.

To solve the above problems, we propose a novel vision transformer based random walk framework for group re-ID. In our method, we design a vision transformer based on a monocular depth estimation algorithm to solve the impact of camera distance on group members by embedding pedestrian depth values into vision transformer. Specifically, we first use monocular depth estimation algorithm to obtain depth maps of single person from the probe image and calculate the average depth of its depth map. Then, the person features are obtained through the vision transformer and constructed into a graph with different nodes according to the size of depth average. It is worth noting that the graph constructed in order of depth value can effectively solve the impact of camera distance on group membership relationships.

In addition, we also design a random walk module to remove members who do not belong to the current group by reconstructing the graph. Specifically, after obtaining graphs with different nodes, we calculate affinity scores between all members in each graph and the gallery images. We then compute the average affinity score for all members in each group, selecting the graph with the highest average affinity score. Subsequently, combining attention mechanisms, we utilize contextual information to propagate new graphs between groups and update graph node features for group matching. It is worth noting that in the process of reconstructing the graph, not only the problem of changes of group members can be effectively solved, but also the problem of group layout changes by group members relative positions can be ignored.

The following are the primary contributions of this paper:
\begin{itemize}
    \item We design a vision transformer based on a monocular depth estimation algorithm to solve the impact of camera distance on group members relationships.
    \item We propose a random walk module by reconstructing the graph to solve the problem of group members and layout changes for group re-ID. 

    \item Our proposed framework is examined on three group re-ID datasets, and experimental results demonstrate that our technique outperforms the most advanced approaches.
\end{itemize}

\section{RELATED WORK}

\begin{figure*}[!t]
\centering
\includegraphics[scale=0.55]{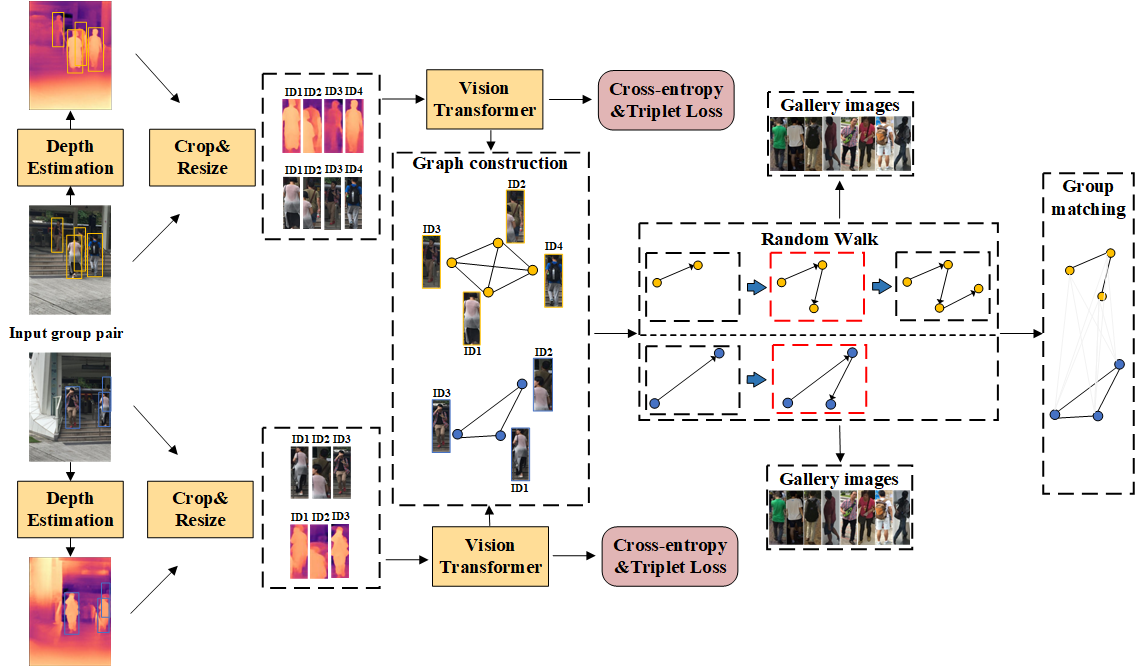}
\caption{Illustration of our proposed framework for group re-identification. First, we take group pairs as input and use the monocular estimation algorithm to obtain the depth map. Second, we crop the single person images and use the vision transformer by embedding the token of person deep values to obtain the single person features. Subsequently, we construct a context graph using person features as nodes. Then, we calculate the affinity scores between all members in each graph and the gallery images. Finally, the node features of the context graph transfer messages with inter-group in the group matching module.}
\label{fig:2}
\end{figure*}

Compared with single person re-ID, there have been relatively few works focusing on group re-ID task \cite{cai2010matching}, \cite{zhu2016consistent}, \cite{zheng2009associating}, \cite{lisanti2017group}, \cite{lin2019group}, \cite{huang2019dot} in the past few years. Some researchers primarily focuses on extracting global or semi-global features to address the challenges posed by the group re-ID. For example, Cai et al. \cite{cai2010matching} presented a covariance descriptor to encode the spatial position and RGB value of each pixel in the group image to capture global features. Zheng et al. \cite{zheng2009associating} suggested CRRRO-BRO descriptor to obtain the global and local features. Due to changes in the relative positions of group members while walking, the group layout has changed, global and local features need to be responsive to such variations. In order to exploit above features in the group, Zhu et al. \cite{zhu2016consistent} presented the method of patch matching to perform similarity matching to measure the distance between the two group images. However, it requires prior constraints on vertical misalignment, which makes it unfeasible in some cases. Xiao et al. \cite{lin2019group} employed multi-granularity information to try to comprehensively capture the features of the group. However, this method is based on traditional manual features and generates an excessive amount of redundant information, resulting in suboptimal accuracy. DotSCN \cite{huang2020dotscn} extracted group consistency features by learning the differential features of paired members in two images. Recently, MACG \cite{yan2020learning} designed complex multi-attention to capture key group features. However, most of the above methods cannot well solve the problem of group members and laylout changes.


\section{Method}
\subsection{Overview}
To address the challenge of group members and layout changes for group re-ID, we design a novel vision transformer based random walk framework, as shown in Fig. \ref{fig:2}. 

In general, 1) our framework can use the monocular estimation algorithm to obtain the depth map and crop the single person images from the probe images and the depth maps respectively; 2) our framework use the vision transformer by embedding the token of person deep values to obtain the person features; 3) We use the person features to construct the graphs in order of depth value; 4) We calculate the affinity scores between all members in each graph and the gallery images to obtain the final graph; 5) The node features of the final graph transfer messages with inter-group in the group matching module.

\subsection{Random Walk}
\label{section:3.2}
Denote ${G_s} = \{ {V_s},{E_s}\}$ as an undirected graph, where $V =\{{v_1},...,{v_n}\}$ represents a vertex set, and $E = \{ {e_1},...,{e_m}\}$ means a set of edges. A square matrix $\bm{W}_{n \times n}$ can be utilized to model a random walk operation on a graph, where $n$ is the number of vertices. The probability of similarity between the $i$-th and $j$-th nodes is represented as $\bm{W}(i, j)$ $\in$ $[0,1]$. In the task of group re-ID, $\bm{W}(i, j)$ can be denoted as the normalized affinity score between the $i$-th and $j$-th person images. Consider $\bm{y}^{(t)}$ be an $n \times 1$-dimensional vector that signifies the affinity scores between the probe image and all gallery images during the $t$-th random walk iteration. Given the matrix $\bm{W}$ containing normalized pairwise affinities among two gallery images, we describe the random walk operation as follows: ${\bm{y}^{(t+1)}} = \bm{W}{\bm{y}^{(t)}}$.

Given a probe image and $n$ gallery images, the Siamese CNN is employed to estimate the pairwise affinity score between images. Ours network processes two images as input and predicts the probability of images belong to the same person. The initial affinity scores generated by CNN between the probe image and gallery images are denoted as $\bm{y}^{(0)}$ $\in$ ${R_n}$. Consider $\bm{S}$ $\in$ ${R_{n\times n}}$ as the matrix that contains the affinity scores between the set of probe sequences and $n$ gallery images. We apply the softmax function to normalize each row of the original affinity matrix $\bm{S}$ to normalize the constraint for all $j$ in $\sum {_j} \bm{W}(i,j) = 1$.

\begin{equation}
\bm{W}(i,j) = \frac{{\exp (\bm{S}(i,j))}}{{\sum\nolimits_{j \ne i} {\exp (\bm{S}(i,j))} }}, {\rm{for \thinspace all \thinspace }}i = 1,...,n,
\label{formulation1}
\end{equation}
where $\bm{W}(i,i)$ = 0 refer to prevent self-reinforcement during random walk iterations. Hence, the diagonal term in Eq~(\ref{formulation1}) does not apply to softmax normalization. An iteration of a random walk on the initial affinity can be represented as:

\begin{equation}
{\bm{y}^{(1)}} = \bm{W}{\bm{y}^{(0)}},
\end{equation}
where $\bm{y}^{(1)}$ represents one iteration of the initial affinity $\bm{y}^{(0)}$. Intuitively, when person $i$ and $j$ share the same ID, their affinities with the probe images should also be similar, and the image groups are more similar. The affinity score $\bm{y}^{(1)}(i)$ of the $i$-th image can be calculated as:

\begin{equation}
{\bm{y}^{(1)}}(i) = \bm{W}(i,1) \cdot {\bm{y}^{(0)}}(1) + ... + \bm{W}(i,n) \cdot {\bm{y}^{(0)}}(n),
\label{formulation2}
\end{equation}

By analogy, we calculate the affinity score $\bm{y}^{(2)}(i)$, ..., $\bm{y}^{(n)}(i)$ of each image in the graph, and obtain the average affinity score of all images. Ultimately, we regard the graph with the highest average affinity as our target group.

\subsection{Group Matching}
Inspired by \cite{yan2020learning}, we perform group matching in graphs by capturing inter-graph information to update graph nodes. In the task of group re-ID, we aim to compute the similarity of two groups. Therefore, it is essential to explore the correlation between groups. In essence, when two groups are the same group, it is likely that the group members should have same correspondences. At the same time, higher similarity of a single pair indicates higher group-level similarity. Given two graphs ($G_s$, $G_r$), we consider the node-level feature pair ($\mathop {\bm{h}}_{si}^{(t - 1)}$, $\mathop {\bm{h}}_{rj}^{(t - 1)}$), where $i$ $\in$ $V_s$ and $j$ $\in$ $V_r$. Furthermore, we divide the features of person $i$ and $j$ into $p$ and $q$ parts respectively. We calculate the importance weight of person features between graphs as:

\begin{equation}
e_{ij}  = \varphi ( \mathbf{W}_e^{(t - 1)}  \bm{h}_{si}^{(t - 1)} , \mathbf{W}_e^{(t - 1)}  \bm{h}_{rj}^{(t - 1)} ),    
\end{equation}
where $\varphi$ is an inner product layer and ${\mathbf{W}^{(t - 1)}_{e}}$ is a projection matrix. Then, we use the softmax function to calculate the attention weights by normalizing the importance weights:

\begin{equation}
\mathop a\nolimits_{ij}  = soft\max (\mathop e\nolimits_{ij} ) = \frac{{\exp (\mathop e\nolimits_{ij} )}}{{\sum\nolimits_k^{(i,k) \in {E_s}} {\exp (\mathop e\nolimits_{ik} )} }}.
\end{equation}

The inter-graph messages with the corresponding attention weights passed from person $i$ in the $p$-th part of the graph $G_s$ to person $j$ in the $q$-th part of the graph $G_r$ can be calculated as follows:

\begin{equation}
\bm{o}_{ij}^{(t)}  = \sum\limits_{i:(i,i) \in {E_s}} { a_{ij} \mathbf{W}_e^{(t - 1)}  \bm{h}_{rjq}^{(t - 1)} }.  
\end{equation}

Here, we focus on person-level similarity when calculating inter-graph attention. As a result, the part-level features within a node are assigned the same set of inter-graph attention weights $\bm{o}_{ij}^{(t)}$. After obtaining the information between graphs, use the fully connected layer to update the node features:

\begin{equation}
\bm{h}_{sip}^{(t)} = MLP( \bm{h}_{sip}^{(t - 1)}, \bm{o}_{ij}^{(t)}).
\end{equation}

The aforementioned feature update steps are iteratively performed for $T$ rounds, utilizing the inter-graph attention mechanism. Subsequently, the model is structured to learn the associations between groups and individuals, respectively. Initially, we create a graph-level representation through readout operations. In this case, we employ self-attention to obtain the ultimated graph representation $\bm{h}_s$ as a weighted sum of node-level features:

\begin{equation}
\bm{u}_i  =  \mathbf{W}_u^{({\rm{T}})} \bm{h}_{si}^{({\rm{T}})},
\end{equation}
\begin{equation}
\gamma_i = \frac{\exp ( \bm{u}_i )}{\sum_{k \in V_s} \exp ( \bm{u}_k )},
\end{equation}
\begin{equation}
\bm{h}_s  = \sum\limits_{i \in  V_s } \gamma_i \mathbf{W}_u^{(\mathrm{T})} \bm{h}_{si}^{(\mathrm{T})}.
\end{equation}

For another graph $G_r$ may be generated $\bm{h}_r$ in the same way. The circle loss function is used to limit the features of the same group and push other groups far apart in order to understand the group correspondence:
\begin{align}
    L_{\rm{g}}^{circle} &= \log [ 1 + \sum\limits_{i = 1}^{K} \sum\limits_{j = 1}^{L} \exp \left( \gamma (a_s^i \bm{h}_s^i - a_r^j \bm{h}_r^j) \right) ] \notag \\
    &= \log [ 1 + \sum\limits_{j = 1}^{L} \exp \left( \gamma a_s^i \bm{h}_s^i \right) \sum\limits_{i = 1}^{K} \exp \left( \gamma a_r^j \bm{h}_r^j \right) ]
\end{align}
where $a_{s}^{j}$ and $a_{r}^{i}$ are non-negative weighting factors, $\gamma$ is as a scale factor. 

\begin{figure}[!t]
\centering
\includegraphics[scale=0.37]{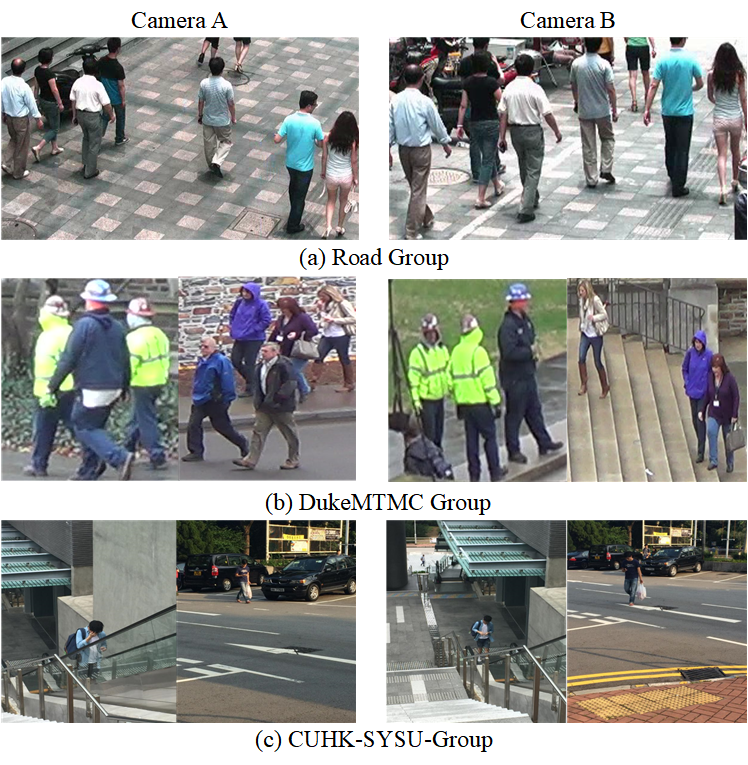}
\caption{Example group images in the (a) Road Group, (b) DukeMTMC Group, and (c) CUHK-SYSU-Group datasets.}
\label{fig:3}
\end{figure}

\begin{table*}[t]
    \caption{In comparison to several group re-ID methods. R-$k$ (k=1, 5, 10, 20) denotes the Rank-k accuracy (\%). The best results are shown are in black boldface font.}

    \renewcommand{\arraystretch}{1}
    \scalebox{1.0}{
    \begin{tabular}{l|cccc|cccc|cccc}
        \hline
        \multirow{2}{*}{\textbf{Methods}} & \multicolumn{4}{c|}{\textbf{CUHK-SYSU-Group}} & \multicolumn{4}{c|}{\textbf{Road Group}}& \multicolumn{4}{c}{\textbf{DukeMTMC Group}} \\ \cline{2-13} 
        ~ & R1 & R5 & R10 & R20 & R1 & R5 & R10 & R20 & R1 & R5 & R10 & R20  \\ \hline
        CRRRO-BRO(BMVC2009) \cite{zheng2009associating} & 10.4 & 25.8 & 37.5 & - & 17.8 & 34.6 & 48.1 & - & 9.9 & 26.1 & 40.2 & - \\ 
        Covariance(ICPR2010) \cite{cai2010matching} & 16.5 & 34.1 & 47.9 & 67.0 & 38.0 & 61.0 & 73.1 & 82.5 & 21.3 & 43.6& 60.4 & 78.2 \\ 
        BSC+CM(ICIP2016) \cite{zhu2016consistent} & 24.6 & 38.5 & 55.1 & 73.8 & 58.6 & 80.6 & 87.4 & 92.1 & 23.1 & 44.3 & 56.4 & 70.4  \\ 
        PREF(CVPR2017) \cite{lisanti2017group} & 19.2 & 36.4 & 51.8 & 70.7 & 43.0 & 68.7 & 77.9 & 85.2 & 22.3 & 44.3 & 58.5 & 74.4  \\ 
        \hline
        LIMI(MM2018) \cite{xiao2018group} & - & - & - & - & 72.3 & 90.6 & 94.1 & - & 47.4 & 68.1 & 77.3 & -  \\ 
        DotGNN(MM2019) \cite{huang2019dot} & - & - & - & - & 74.1 & 90.1 & 92.6 & - & 53.4 & 72.7 & 80.7 & -  \\ 
        GCGNN(TMM2020) \cite{zhu2020group} & - & - & - & - & 81.7 & 94.3 & 96.5 & 97.8 & 53.6 & 77.0 & 91.4 & 94.8  \\ 
        MACG(TPAMI2020) \cite{yan2020learning} & 63.2 & 75.4 & 79.7 & 84.4 & 84.5 & 95.0 & \textbf{96.9} & 98.1 & 57.4 & 79.0 & 90.3 & 94.3  \\
        MGR(TCYB2021) \cite{lin2019group} & 57.8 & 71.6 & 76.5 & 82.3 & 80.2 & 93.8 & 96.3 & 97.5 & 48.4 & 75.2 & 89.9 & 94.4  \\ 
        DotSCN(TCSVT2021) \cite{huang2020dotscn} & - & - & - & - & 84.0 & 95.1 & 96.3 & \textbf{98.8} & \textbf{86.4} & \textbf{98.8} & \textbf{98.8} & \textbf{98.8}  \\ 
         \hline
        Ours & \textbf{80.1} & \textbf{89.5} & \textbf{92.3} & \textbf{95.2} & \textbf{85.6} & \textbf{95.8} & 96.7 & 97.4 & 71.8 & 78.3 & 83.2 & 88.4  \\  
        \hline
    \end{tabular}
    }
    \label{table:2}
\end{table*}


\section{EXPERIMENTAL RESULTS}
\subsection{Datasets and Experimental Settings}

We evaluate our proposed group re-ID method on three publicly available datasets: (1) the Road Group dataset (RG) contains 324 images including 162 group classes, (2) the DukeMTMC Group dataset (DG) contains 354 images including 177 group classes captured by 8 cameras, and (3) CUHK-SYSU-Group dataset (CSG) based on CUHK-SYSU dataset contains 3,839 images including 1,558 group classes. Examples of group images from three datasets are illustrated in Fig.~\ref{fig:3}.

We randomly split datasets in half to create training and testing sets, using vision transformer as our backbone. We resize the person images as 256×128 for inputs. The initial learning rate is set at 0.0001 trained to 300-th epoch. For simplicity, graphs are constructed with an equal number of empty nodes for groups of varying sizes, and dummy nodes are added for groups with limited members. 



\subsection{Compared with Other Group Re-ID Methods}
We assess the effectiveness of the proposed approach against existing methods using the RG, DG, and CSG datasets, as shown in Table~\ref{table:2}. The current methods are categorized into two groups: handcrafted methods and deep learning methods. It is worth noting that the deep learning method DotSCN incorporates additional datasets for auxiliary training. Among these, MACG is considered the top-performing method for single dataset training, while DotSCN excels in multiple dataset training scenarios. The results demonstrate that our proposed method in this paper achieves the advanced performance in single dataset training. Compare with MACG, our method exceeds 1.1\%, 14.4\% and 16.9\% Rank 1 on RG, DG and CSG datasets. Even without the use of additional datasets, our approach outperforms DotCSN in certain instances, underscoring the superiority of our method.


\begin{figure}[!t]
\centering
\includegraphics[scale=0.53]{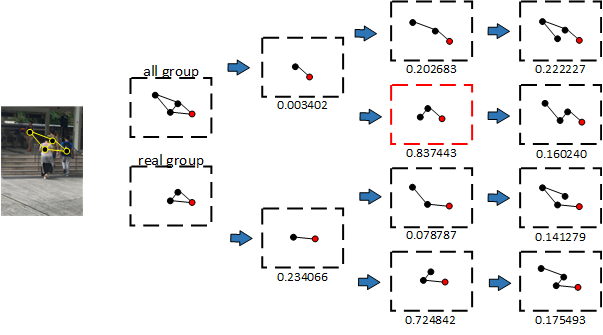}
\caption{The visualization of the random walk module process.}
\label{fig:5}
\end{figure}

\subsection{The visualization of results}

We randomly select a probe image in the CSG dataset to visualize the process of the random walk module. Each image has a ground truth group as the criterion for the affinity score. We first build the image into a graph structure according to the average size of the depth values, and then use different vertices as starting nodes to construct the graph structure through the random walk method, as shown in the Fig~\ref{fig:5}. The numbers represent the affinity score of each graph structure with the ground truth graph structure. We find that the group member with the highest affinity score happens to be the ground truth group member in the graph. Experiments have shown that our method can prove that groups composed of pedestrians whose depth values are more similar are more likely to be real groups. At the same time, this method can also effectively remove pedestrians who do not belong to the group and ignore the group layout changes.

\begin{figure}[!t]
\centering
\includegraphics[scale=0.53]{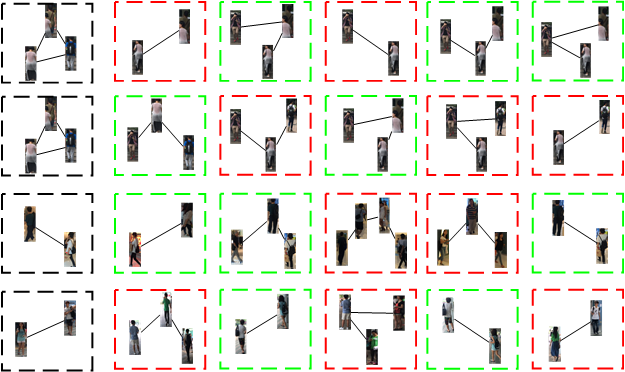}
\caption{The visualize results of the top-5 ranking lists with our method on CSG dataset. Images with green and red borders indicate correct and incorrect matches, respectively.}
\label{fig:9}
\end{figure}

\textbf{Group Matching} The Fig.~\ref{fig:9} shows the matching results of our proposed method. The first two examples show situations which group members changes between queries and gallery images. We observe that even if there is occlusion in these examples, the results can still be retrieved correctly. The last two examples show failing situations where the gallery group typically contains people who share a similar appearance to the two person in the query images. In this case, it will be more difficult for the model to retrieve the correct matches.

\subsection{Ablation Study}

\textbf{Body Parts Ablation.} We evaluate the effects of various human body. To be specific, we segment them into $P$ parts, and results of various partition configurations are shown in Table~\ref{table:33}. It is evident from the results that a $P$ = 2 partition yields notably lower performance compared to other partitioning schemes, primarily due to its coarse segmentation. Conversely, the most favorable performance is attained when employing $P$ = 4 partitions.

\begin{table}[!t] 
\caption{Body parts ablation study of our proposed method on the CSG datasets.}
\newcolumntype{C}{>{\centering\arraybackslash}X}
\begin{tabularx}{\linewidth}{lCCC}
\toprule
~ & \textbf{Rank 1} &\textbf{Rank 5} &\textbf{Rank 10}  \\
\midrule
    $P$ = 2 & 72.1 & 79.1 & 84.2    \\ 
    $P$ = 4 & 80.1 & 89.5 & 92.3   \\ 
    $P$ = 6 & 78.5 & 88.3 & 91.2   \\ 
    $P$ = 8 & 77.4 & 87.8 & 90.8   \\ 
\bottomrule
\end{tabularx}
\label{table:33}
\end{table}

\textbf{Backbone Networks Ablation.} We analyze how different backbone networks affect model performance. We use ResNet50, DensNet161, EfficientNet-b7 and Vision transformer as backbone networks respectively. Table~\ref{table:3} shows the experimental results. We find that using Vision transformer as the backbone network has the best performance, while using EfficientNet-b7 as the backbone network has the worst performance.

\begin{table}[!t] 
\caption{Backbone networks ablation study of our proposed method on the CSG datasets.}
\newcolumntype{C}{>{\centering\arraybackslash}X}
\begin{tabularx}{\linewidth}{lCCC}
\toprule
~ & \textbf{Rank 1} &\textbf{Rank 5} &\textbf{Rank 10}  \\
\midrule
    ResNet50 & 68.3 & 74.1 & 80.2    \\ 
    DensNet161 & 65.6 & 70.2 & 78.8   \\ 
    EfficientNet-b7 & 61.9 & 67.7 & 72.5   \\ 
    Swin transformer & 75.2 & 82.5 & 88.7   \\ 
    Vision transformer & 80.1 & 89.5 & 92.3    \\
\bottomrule
\end{tabularx}
\label{table:3}
\end{table}


\textbf{Module Ablation.} We conduct module ablation experiments on CSG dataset to confirm the effectiveness of our proposed module. Table~\ref{table:11} shows the experimental results. We examine distinct variants of the framework such as Random Walk module (RW), Group Matching module (GM) and Circle Loss (CL) function. By rearranging elements in different arrangements, we can determine the final model performance.



\begin{table}[!t]
    \centering
    \tabcolsep=0.15cm
    \renewcommand{\arraystretch}{1.5}
    \caption{Modules ablation study of our proposed method on the CSG dataset.}

    \scalebox{0.75}{\begin{tabular}{l|ccccccc}
    \hline
        Modules & Base & Variants  & Variants & Variants & Variants & Variants & Variants \\ \hline
        RW &    & \Checkmark &   &   & \Checkmark &   & \Checkmark\\
        GM &    &   &  \Checkmark &   &   &  \Checkmark & \Checkmark \\ 
        CL &    &   &   &  \Checkmark &  \Checkmark &  \Checkmark &\Checkmark \\ \hline
        Rank 1 & 63.2 & 73.2  & 69.2  & 71.3  & 77.8  & 73.9  & 80.1 \\ \hline
    \end{tabular}}
    \label{table:11}
\end{table}

\section{Conclusion}
In this paper, we present a novel vision transformer based random walk framework to address the challenge of group re-ID, which contain a vision transformer based on a monocular depth estimation algorithm to construct a graph through the average depth value of pedestrian features to fully consider the impact of camera distance on group members relationships and a random walk module to reconstruct the graph by calculating affinity scores between target and gallery images to remove pedestrians who do not belong to the current group. Experiments show that we obtained outstanding results on three group re-ID datasets, demonstrating the efficacy of our suggested methodology. 


\bibliographystyle{IEEEbib}

\end{document}